\documentclass[10pt,twocolumn,letterpaper]{article}

\usepackage[pagenumbers]{wacv} 

\usepackage{graphicx}
\usepackage{amsmath}
\usepackage{amssymb}
\usepackage{booktabs}
\usepackage{placeins}
\usepackage{multirow}
\usepackage{multicol}
\usepackage{bbding}
\usepackage{pifont}
\usepackage{enumitem}
\usepackage{tabularx}
\usepackage[accsupp]{axessibility}
\usepackage[pagebackref,breaklinks,colorlinks]{hyperref}

\usepackage[capitalize]{cleveref}
\crefname{section}{Sec.}{Secs.}
\Crefname{section}{Section}{Sections}
\Crefname{table}{Table}{Tables}
\crefname{table}{Tab.}{Tabs.}

\def\wacvPaperID{478} 
\def\confName{WACV}
\def\confYear{2024}

\begin{document}
\title{Salient Object Detection for Images Taken by People With Vision Impairments} 
\author{Jarek Reynolds*, Chandra Kanth Nagesh*, and Danna Gurari\\
* denotes equal contribution\\
University of Colorado Boulder\\
}
\maketitle

\begin{abstract}
Salient object detection is the task of producing a binary mask for an image that deciphers which pixels belong to the foreground object versus background. We introduce a new salient object detection dataset using images taken by people who are visually impaired who were seeking to better understand their surroundings, which we call VizWiz-SalientObject. Compared to seven existing datasets, VizWiz-SalientObject is the largest (i.e., 32,000 human-annotated images) and contains unique characteristics including a higher prevalence of text in the salient objects (i.e., in 68\% of images) and salient objects that occupy a larger ratio of the images (i.e., on average, $\sim$50\% coverage).  We benchmarked ten modern models on our dataset.  One method achieves nearly human performance while the rest struggle, mostly for images with salient objects that are large, have less complex boundaries, and lack text as well as for lower quality images. To facilitate future extensions, we share the dataset at \href{https://vizwiz.org/tasks-and-datasets/salient-object-detection}{https://vizwiz.org/tasks-and-datasets/salient-object-detection}.
\end{abstract}
\section{Introduction}
\label{sec:intro}
Locating the most prominent foreground object in an image is a core computer vision problem, often referred to as salient object detection (as well as salient object segmentation and foreground object detection/segmentation) \cite{looktel,taptapsee,cheng2014global,borji2015salient}.  Our motivation is to have salient object detection models work well for images taken by people who are blind or who have low vision\footnote{People with low vision have limited visual abilities that cannot be improved by wearing glasses or surgery.}. Such a feature would offer several benefits to this community.  Users of mobile phone applications which describe one's visual surroundings (with help from remote humans or algorithms)~\cite{Aipoly:online, Aira:online, BeMyEyes:online, BeSpecul:online, iDentifi:online, desmond_microsofts_nodate, looktel, taptapsee} could use salient object detection to support privacy-preservation and impression management~\cite{alharbi2022understanding}.  In particular, users could obfuscate all content except the foreground content of interest to avoid sharing information inadvertently captured in the background of images~\cite{gurari2019vizwiz}.\footnote{Some organizations record submitted data for the potential that it could be useful evidence in legal contexts. }  Users of such applications also could use it to obfuscate background content so that visual descriptions focus only on the content of interest, the salient object~\cite{alharbi2022understanding}.  Individuals also could edit their photos to crop the salient content, including for sharing on social media~\cite{voykinska2016blind}.  Finally, low vision individuals could use it to quickly magnify the content of interest, the foreground object, for deeper inspection~\cite{afb_magnifiers,stangl2018browsewithme}.  


\begin{figure}[t!]
    \centering
    \includegraphics[width=24em]{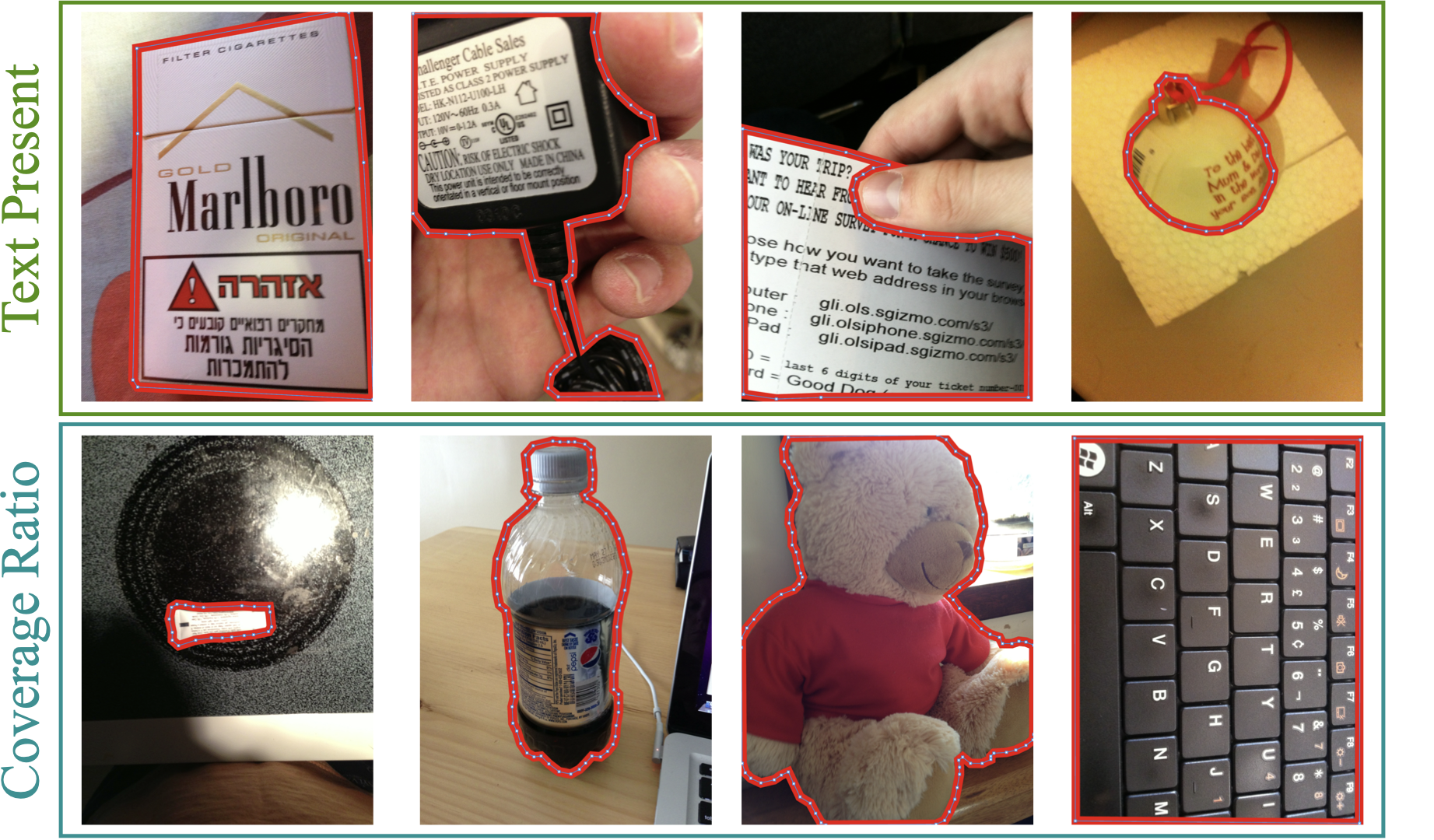}
    \caption{Images exemplifying unique aspects of our new VizWiz-SalientObject dataset when compared to other datasets.  The salient objects commonly contain text and occupy a larger portion of the image (i.e., high coverage).}
    \label{fig:final_image_merge}
\end{figure}

While many salient object detection datasets have been created to enable progress in algorithm development~\cite{borji2015salient,borji2019salient,gupta2020,Wang_2019_CVPR}, they are typically built using high-quality images collected from photo-sharing websites on the Internet.  Yet, it has been shown that images taken by visually impaired photographers who are trying to learn about the content they photograph~\cite{gurari2018vizwiz,gurari2019vizwiz} often have distinct characteristics from images in mainstream computer vision datasets, including different types of content such as objects showing text~\cite{gurari2020captioning} and many quality issues since photographers cannot verify the quality of images they take~\cite{chiu2020assessing}.  This begs a question of how well algorithms designed for and trained on existing datasets will generalize to the real-world challenges observed in images taken by people with vision impairments.  

To fill this gap, we introduce a new salient object detection dataset based on images captured in an authentic use case where visually impaired photographers shared their images to solicit assistance in learning about the visual content.  In particular, we crowdsourced the collection of salient object annotations to create a final dataset consisting of 32,000 annotated images.  Examples of annotated images are shown in Figure~\ref{fig:final_image_merge}.  We call our dataset VizWiz-SalientObject (or VizWiz-SO).

We also conduct a detailed analysis to reveal how this new dataset relates to existing datasets.  When compared to seven existing salient object detection datasets, we observe VizWiz-SalientObject is the largest (i.e., 32,000 human-annotated images) and is unique in its higher prevalence of salient objects containing text (i.e., in 68\% of images),  occupying a larger ratio of the images (i.e., on average, $\sim$50\%), and having less complex boundaries. When comparing our salient objects to the visual evidence needed to answer questions the photographers asked about their images (i.e., taken from the VizWiz-VQA-Grounding dataset~\cite{chen2022grounding}), we observe that over half the images have a salient object that matches the necessary visual evidence.  

We then benchmark ten modern models on our dataset. Experiments reveal that they struggle for lower quality images as well as for images with salient objects that are large, have less complex boundaries, and lack text. To our knowledge, this work is the first to reveal these limitations of algorithms, and we attribute these findings in part to the distinct visual characteristics of our new dataset. Excitingly, one model nearly achieves human performance. 

We expect this new dataset will support developing more generalized models that can also benefit related applications with similar real-world challenges (e.g., lower quality images, large objects, little object boundary complexity), including robotics, lifelogging, and other egocentric-based applications.  We also anticipate that algorithmic improvements will benefit downstream applications from salient object detection, such as salient object tracking \cite{10045751,9711449, 8954050}, visual question answering, and image captioning.  To facilitate extensions, we publicly-share it at \href{https://vizwiz.org/tasks-and-datasets/salient-object-detection}{https://vizwiz.org/tasks-and-datasets/salient-object-detection}.




\section{Related Work} 
\label{rel}
\paragraph{Salient Object Detection Datasets.}
Over the past couple of decades, many datasets were introduced to facilitate improving the design of algorithms that address salient object detection problems. Several survey papers provide comprehensive characterizations of the tens of datasets designed for this task~\cite{borji2015salient,borji2019salient,gupta2020,Wang_2019_CVPR}.  A common observation is that datasets were artificially constructed around high-quality images which often feature salient objects in the center of the images with a high contrast against the background.  This is a mismatch from many real-world settings, especially for visual media taken by visually impaired photographers who often photograph distinct types of content with the aim to learn about that content.  We introduce the first salient object detection dataset based on images taken by visually impaired people in an authentic use case where they were trying to learn about their visual surroundings. We show that, compared to seven modern datasets, our dataset is larger and has a high prevalence of salient objects containing textual information, occupying larger portions of the images, and having little boundary complexity.

\vspace{-1em}\paragraph{Salient Object Detection Models.}
Novel models for performing salient object detection have been introduced for over 20 years, with the status quo since 2015 being that state-of-the-art methods employ neural networks trained on large-scale annotated datasets.  Several survey papers provide comprehensive characterizations of the many models for this task~\cite{borji2015salient,borji2019salient,gupta2020,Wang_2019_CVPR}.  While convolutional neural network (CNN) based models became the mainstream method \cite{u2net, pfsnet, f3net} in 2015, transformer based models \cite{pgnet, vst} have become the mainstream approach over the past few years.  To assess how well modern methods perform on our new dataset, we benchmark ten modern methods.  While most methods fall below human performance, struggling for lower quality images as well as for images with salient objects that are large, have little boundary complexity, and lack text, one method nearly achieves human performance.
 

\vspace{-1em}\paragraph{Visual Assistance Applications.}
Automated salient object detection methods are desired by people with vision impairments to independently achieve many tasks. Examples include for privacy preservation and impression management~\cite{alharbi2022understanding}, by obfuscating background content in images (or video frames) to avoid sharing personal information inadvertently~\cite{gurari2019vizwiz}. This could benefit the 100s of thousands of users who submit millions of requests to numerous technologies~\cite{Aipoly:online, Aira:online, BeMyEyes:online, BeSpecul:online, iDentifi:online, desmond_microsofts_nodate, looktel, taptapsee} to receive visual assistance for daily tasks such as deciding what to eat, wear, and buy~\cite{brady2013investigating,gurari2019vizwiz}.  Such models would also support more efficient learning about the salient content in images by restricting image descriptions to focus only on that content~\cite{alharbi2022understanding} or rapidly magnifying that content for low vision users' inspection~\cite{afb_magnifiers,stangl2018browsewithme}.  Finally, such models would enable users to independently edit content in their images~\cite{voykinska2016blind} and decide what to share to social media~\cite{zhao2017effect}.
\section{VizWiz-SalientObject Dataset}
\label{dataset-creation}
We now introduce our new salient object detection dataset, we call VizWiz-SalientObject (VizWiz-SO).

\subsection{Dataset Creation}
\label{dataset-creation-process}
\paragraph{Image Source.}
We focus on images taken by visually impaired people who shared them in an authentic use case where they were soliciting visual assistance.  Specifically, we leverage the 39,181 labeled images from the VizWiz-Captions dataset, each of which is paired with five crowdsourced captions~\cite{gurari2020captioning}.  Observing that images from these photographers can have severe quality issues resulting in no detectable salient object (e.g., extreme blur or inadequate illumination), we did not use the images which were captioned as follows by at least four of the five crowdworkers: ``Quality issues are too severe to recognize visual content."  We also did not use the small images (i.e., both the height and width were less than 300 pixels) because of the challenges of collecting precise annotations for such images.  This left us with 37,120 images for our annotation task.

\vspace{-1em}\paragraph{Task Design.}
Our task interface for segmenting salient objects begins with a comprehensive instruction set at the top detailing both how to navigate the interface and how to complete challenging annotation scenarios.  Next, it shows an image alongside two preliminary questions for verifying there is a single, unambiguous foreground object.  The first question asks ``Is the image showing a screenshot?"  If the answer is ``yes", we conclude the image lacks a salient object.  Next, we ask the more general, direct question of ``Is there a single unambiguous foreground object?"  An annotator is only prompted to segment the foreground object for images deemed by these preliminary questions to show a single, unambiguous foreground object.  

To demarcate the boundary of the salient object, the interface collects a series of points that are connected into polygon(s).  When segmenting the salient object, the annotator is required to remove any \emph{holes} (e.g., donut) as well as capture all object parts when \emph{occlusions} break a salient object into more than one polygon (e.g., hand obfuscates a pencil into two parts).  The annotator also has an option to select a button indicating that the salient object occupies the full image.  

\vspace{-1em}\paragraph{Annotation Collection.}
We leveraged the benefits of an around-the-clock distributed workforce by crowdsourcing annotations via Amazon's crowdsourcing marketplace, Amazon Mechanical Turk (AMT).  

Although AMT can support our large-scale annotation needs, it brings concerns about annotation quality due to the anonymous nature of the workforce. Consequently, we implemented several measures to ensure the collection of high-quality annotations.  First, we only accepted workers who had at least a 98\% acceptance rate and had completed at least 500 Human Intelligence Tasks (HITs) on AMT.  To encourage understanding of our task instructions, we also only accepted crowdworkers from the United States since that gave us confidence that they have English-language proficiency. We also required crowdworkers to pass a qualification test covering five challenging annotation scenarios, including annotating foreground objects with complex boundaries, holes, and occlusions.  

We employed 40 AMT crowdworkers who completed our qualification test to annotate all images.  For each of the 37,120 images, we collected two annotations from the crowdworkers.\footnote{For a subset of images, we collected four annotations to support further analysis of human performance (see the Supplementary Materials).}  During annotation collection, we monitored ongoing quality by tracking each worker's performance with respect to their frequency of indicating the presence of full-screen annotations or no prominent foreground object as well as the level of detail they provided in their segmentations (e.g., high prevalence of triangles).  Cumulatively, the crowdworkers took 1,290 annotation hours over 11 days to complete annotating the 37,120 images. 

\vspace{-1em}\paragraph{Annotation Post-Processing.}
We next analyzed the redundant annotations per image to determine how to use each annotated image in the final dataset.  First, we removed 3,662 images for which workers agreed there was no single, unambiguous salient object, which occurred when both annotators either answered ``Yes" to ``Is the image a screenshot?" or ``No" to ``Is there a single most prominent foreground object?"  Next, we manually inspected 7,443 images for which workers disagreed on the answers to either of the two preliminary questions and determined whether there is indeed a single, unambiguous object.  Finally, with all images deemed to have a single, unambiguous salient object, we determined which annotation to assign as ground truth.  To assist in this process, we computed the intersection over union (IoU) score between the two segmentations for all images with two or more segmentations.  With IoUs \begin{math}\geq\end{math} 0.90, we deemed both annotations high quality and randomly selected one as ground truth.  For the remaining 2,951 images with IoUs $< 0.90$, we manually reviewed the annotations to decide whether one was correct or whether the image should be discarded due to foreground object ambiguity.

\subsection{Dataset Analysis}
\label{dataset_analysis}
We now characterize the VizWiz-SalientObject dataset (VizWiz-SO) and how it relates to existing datasets.

\begin{table*}[t!]
\centering
\begin{tabular}{lcccccccc}
\hline
 & DAVIS-S \cite{zeng2019towards} & PASCAL-S \cite{pascal-s} & HR \cite{zeng2019towards} & ECSSD \cite{ecssd} & DUT-O \cite{duts-omron} & UH \cite{pgnet} & DUTS \cite{wang2017} & Ours \\ \hline\hline
Images & 92 & 850 & 2,010 & 1,000 & 5,168 & 5,920 & 15,572 & \textbf{32,000} \\ \hline
Text  & 13\% & 24\% & 15\% & 15\% & 11\% & 19\% & 13\% & \textbf{68\%} \\ \hline
MR  & 22\% & 31\% & 25\%  & 9\%  & 17\%  & \textbf{35\%}  & 19\%  & 1\%  \\ \hline
Holes  & \textbf{82\%} & 50\% & 62\% & 29\% & 28\% & 75\% & 41\% & 4\% \\ \hline
MW & 1080 & 375 & 2704 & 300 & 300 & 3612 & 300 & 1296 \\ \hline
MH & 1920 & 500 & 3264 & 400 & 400 & 5000 & 400 & 968 \\ \hline
\end{tabular}
\centering
\vspace{0.1em}
\caption{Characterization of our VizWiz-SO dataset and seven existing salient object detection datasets with respect to how many images are included (``Images"), percentage of images containing text in the salient objects (``Text"), percentage of images that have salient objects consisting of more than one region (``MR"), percentage of images that have holes in the salient objects (``Holes"), median image width (``MW") and median image height (``MH").  As shown, our dataset is distinct in that it contains more images, more salient objects with text present, more salient objects consisting of one region, and fewer salient objects containing holes.  (HR=HRSOD; UH=UHRSD) }
\label{tab:simple_comp}
\end{table*}

\vspace{-0.2em} \subsubsection{VizWiz-SalientObject vs Existing SOD Datasets}
\label{sec_comparisonToOtherDatasets}
We first characterize our new dataset and how it is both distinct and similar to the following seven SOD datasets:

\begin{itemize}[itemsep=0.1em,leftmargin=*]
\item \textit{DUTS} \cite{wang2017}: commonly used to train state-of-the-art algorithms (e.g., \cite{vst, pgnet, u2net, basnet, f3net, pfsnet}) due to its large size paired with diverse saliency challenges. 

\item \textit{DUT-OMRON} \cite{duts-omron}: introduced in 2013, this older dataset was popular for a while.

\item \textit{ECSSD} \cite{ecssd}: designed to show complex scenes with textures and structures expected to be common in real-world salient object detection scenarios. 

\item \textit{PASCAL-S} \cite{pascal-s}: derived from PASCAL VOC's \cite{pascal-voc} validation set to facilitate salient object segmentation on a popular set of images in the computer vision community. 

\item \textit{HRSOD} \cite{zeng2019towards}: explicitly designed for salient object detection on high-resolution images. This is relevant for our use case since images taken by people with vision impairments often can be high resolution.

\item \textit{UHRSD} \cite{pgnet}: currently the largest ultra-high resolution salient object detection dataset, further supporting our use case since images can be ultra-high resolution. 

\item \textit{DAVIS-S} \cite{zeng2019towards}: derived from DAVIS \cite{Perazzi_CVPR_2016}, a densely annotated video segmentation dataset.  This is relevant for our use case to analyze implications for video frames since visually impaired photographers often stream live video when using visual assistance technologies~\cite{Aira:online, BeMyEyes:online}.

\end{itemize}

\noindent
Of note, images in six of these datasets originate from the Internet on photo-sharing websites such as Flickr \cite{wang2017, duts-omron, ecssd, pascal-s, zeng2019towards, pgnet}, and so likely are high quality since they were deemed of sufficient quality to upload to the Internet.\footnote{The origins of the images for \cite{zeng2019towards} is not reported.} 

For each salient object in every dataset, we characterize it in eight ways.  Three measures detect the presence of particular properties for the salient object.  These are whether the salient object \textit{contains text}~\footnote{We obfuscate all image content but the salient object and then check whether Microsoft Azure's OCR API returns text.}, \textit{consists of multiple regions}~\footnote{Multiple regions means there are multiple separate polygons.  This can occur either because multiple salient objects were annotated or because of occlusions that lead to more than one region for a single salient object.}, or \textit{contains any hole(s)}.  The next three measures characterize the salient region itself.  First, we identify the position of an object within an image by measuring its \textit{center of mass} relative to the image coordinates, resulting in $x$ and $y$ coordinate values in the range between 0 to 1.  Next, we characterize the object's \textit{boundary complexity} by computing its isoperimetric inequality, which is the ratio of the object's area to the length of its perimeter. Values range from 0 to 1, with larger values indicating simpler boundaries that are less jagged/dented (e.g., a circle).  Finally, to gauge the relative size of a salient object in the image, we compute its \textit{coverage ratio}, meaning the fraction of all image pixels that are occupied by the salient object's pixels. The final two measures offer finer-grained details about typical image resolutions.  We report the \textit{median image width} and \textit{median image height}.

\begin{figure*}[t!]
    \centering
    \includegraphics[width=1\textwidth]{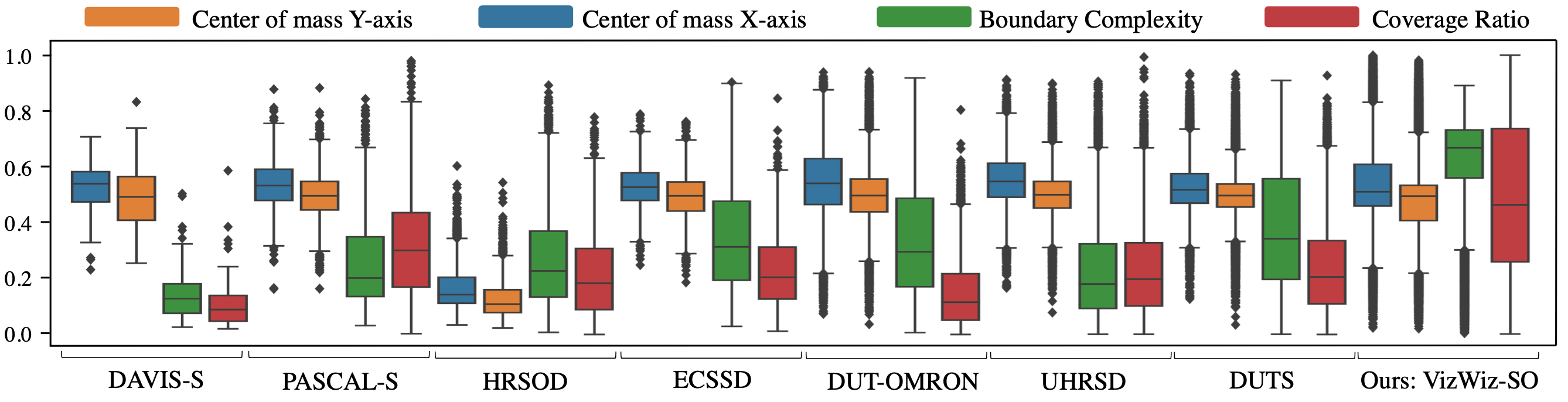}
    \caption{Summary statistics for ours and seven other datasets with respect to four measures.  Each box reveals statistics about all salient objects in a particular dataset with the central mark capturing the median value, box edges the 25th and 75th percentiles values, whiskers the most extreme data points not considered outliers, and individually plotted points the outliers.  Our dataset is unique in that salient objects tend to have less complex boundaries, occupy larger portions of an image, and exhibit a greater diversity of sizes relative to the image.}
    \label{fig:final_barplot}
\end{figure*}

We show summative statistics of our findings per dataset in Table \ref{tab:simple_comp} and Figure \ref{fig:final_barplot}.  In Table \ref{tab:simple_comp}, we report how many images are in each dataset paired with the percentage having salient objects with text, multiple regions, and holes as well as summative statistics about image resolutions.  \emph{To our knowledge, this is the first paper that systematically characterizes SOD datasets based on whether text is present, holes are present, and multiple regions are present.}  In Figure~\ref{fig:final_barplot}, we visualize statistics summarizing the values for each dataset's salient objects with respect to center of mass, boundary complexity, and coverage ratio using boxplots.  

A unique aspect of our VizWiz-SO dataset is that it features more salient objects with textual data. Specifically, 68\% of salient objects in VizWiz-SO contain text while the dataset with the next highest prevalence of text, PASCAL-S \cite{pascal-s}, only has it for 24\% of the images (Table \ref{tab:simple_comp}). A gap of this magnitude (i.e., 44 percentage points) suggests that our new dataset offers a considerable domain shift in the salient object detection problem space.  We suspect part of this shift stems from the types of salient objects included, with many more daily objects such as products (e.g., food packages)  included in our VizWiz-SO dataset.

Another unique aspect of VizWiz-SO is that far fewer images feature salient objects that consist of multiple regions; i.e., only 1\% of images (Table \ref{tab:simple_comp}).  We suspect this distinction stems from our unique approach of adopting a rigorous annotation preprocessing step, where we require crowdworkers to verify images have one unambiguous salient object before allowing them to annotate images for use in our final dataset.  Any remaining objects in our dataset with multiple regions are therefore highly likely a result of occlusions breaking a single salient object into multiple polygons, which evidently is incredibly rare.

VizWiz-SO is also unique due to the rarity in which salient objects contain holes; i.e., only observed for 4\% of images (Table \ref{tab:simple_comp}).  From visual inspection, we suspect this finding reflects a domain shift in the types of content found in the datasets.  For example, objects in other datasets with holes include people riding bikes, people dancing, and animals in intricate poses. In VizWiz-SO, objects with holes include retail packaging made to hang from hooks, pairs of scissors, and coffee mugs.  

A further distinction of our VizWiz-SO dataset is that the salient objects tend to have less complex boundaries (Figure~\ref{fig:final_barplot}).  We suspect this is again because of a domain shift in the types of objects in our dataset, with many more human-made items, such as food packaging boxes and cans, that by design have simpler shapes.  

A final distinction of salient objects in our VizWiz-SO is how much of the image they occupy (Figure~\ref{fig:final_barplot}).  First, they tend to occupy a much larger amount of the image than observed in other datasets.  Specifically, they on average occupy roughly half of all image pixels, with a mean coverage ratio of 0.5 and a median of 0.46.  In contrast, the dataset with the next highest coverage ratio statistics is PASCAL-S~\cite{pascal-s}, and over 75\% of its images contain salient objects that occupy less than half of the image pixels.  We attribute this distinction to the authentic use case of our dataset, where visually impaired photographers attempting to learn about the salient objects they are photographing seem to be taking zoomed-in or close-to-camera images of the content of interest.  Another unique aspect of our salient objects, is that they exhibit a larger range of sizes, as shown by the gaps between the 25 and 75 percentile values of each box. For example, PASCAL-S features the next largest interquartile range with a 23\% gap(i.e., 19\% to 42\%).  In contrast, the gap for VizWiz-SO is more than twice as large at 56\% (i.e., 22\% to 78\%).  Consequently, a unique challenge of our dataset for algorithms is that they no longer can assume a strong bias regarding a salient object's relative size.

While our findings highlight that our VizWiz-SO dataset has many distinct characteristics, one commonality it has with most existing salient object detection datasets that we found surprising is that the salient objects typically occupy centered positions within an image.  We observe this trend for all datasets except HRSOD in Figure~\ref{fig:final_barplot}. We found this surprising since visually impaired photographers cannot visually inspect their images to verify image quality.  This finding suggests that visually impaired photographers have skills in conforming to the common photographer's bias of centering contents of interest they are trying to photograph.

\subsubsection{Salient Objects vs Answer Groundings for Visual Question Ansswering (VQA)}
\label{sec:SOvsVQA}
\begin{figure}[t!]
    \begin{center}
    \includegraphics[width=24em]{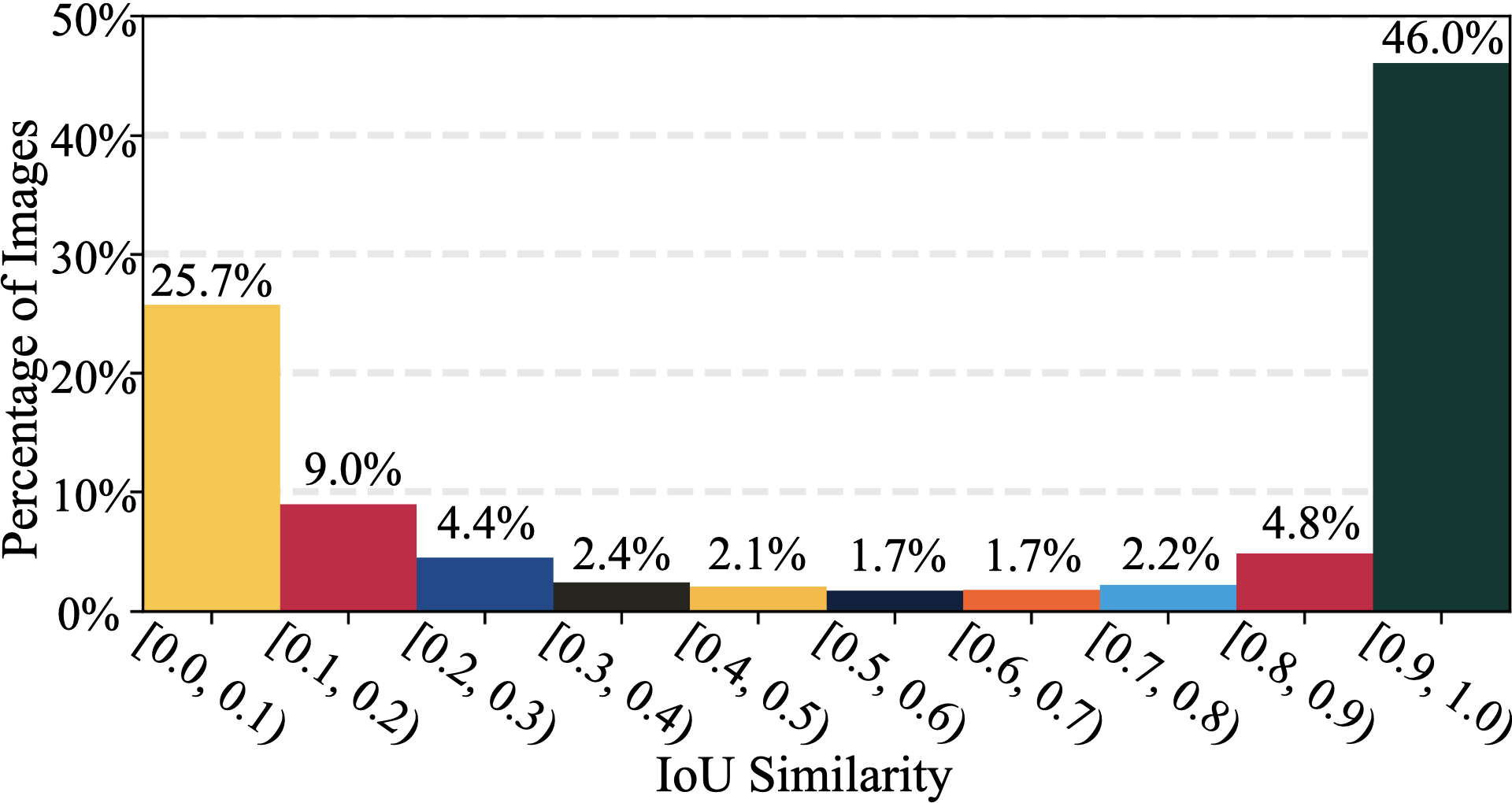}
    \caption{Frequency of observing different levels of similarity between the salient object and the visual evidence needed to answer the photographer's question. As shown, the photographers often wanted to learn about the salient objects in their images.} 
        \label{fig:salientObject-vs-answerGrounding}
    \end{center}
\end{figure}

We next explore how the target content the photographers were asking about relates to an image's salient object.  To do so, we compare the annotations of the visual evidence needed to answer questions about the images, i.e., \emph{answer groundings} provided in the VizWiz-VQA-Grounding dataset \cite{chen2022grounding}, to annotations of the salient objects.  We first identified all images that were in common across the two datasets, yielding 6,540 images. For each image, we then measured the similarity between the answer grounding and salient object segmentations using the IoU metric.  We visualize our results using a histogram where we categorize each image into one of ten interval bins starting with IoU=[0.0, 0.1), incrementing in intervals of 0.1, and ending with IoU=[0.9, 1.0).  Results are shown in Figure \ref{fig:salientObject-vs-answerGrounding}.

Roughly half of the images have a high similarity between the salient object and answer grounding; e.g., 46\% have an IoU \begin{math}\geq\end{math} 0.9.  This reveals that visually impaired photographers often tried to learn about the salient object when trying to get answers to their visual questions.  This reinforces prior works' findings that users of visual assistance technologies most commonly trying to learn about a (salient) object~\cite{brady2013investigating,brady2015crowdsourcing,zeng2020vision,gurari2018predicting,kacorri2017people}.

We also observe that roughly one quarter of the images have a very low similarity between the salient object and answer grounding; i.e., 25.7\% of images had an IoU \begin{math}<\end{math} 0.1.  From manual review of 1,680 images with IoUs less than 0.1, we discovered that 95\% (i.e., 1,599) have a salient object featuring a full-screen or large region while the answer grounding captures a small aspect of the salient object. Examples include expiration dates on food packages or the current page number of an open book. The remaining 5\% (i.e., 81) of these images featured a answer grounding unrelated to the salient object.

More generally, we observe that the IoU scores follow a U-shaped distribution with only a small portion of images having middling scores; e.g., 7.9\% (i.e., 511) have an IoU \begin{math}\geq\end{math} 0.3 and \begin{math}<\end{math} 0.7. Among these, the salient object always contained the answer grounding region. Two primary trends led to these less common IoU scores. The first is that larger answer grounding regions occur with smaller salient objects. Examples include brands of cereal, types of soda, and denominations of currency. The second trend was for salient objects featuring holes. The VizWiz-VQA-Grounding dataset did not account for holes in their annotation task, and this often led to lower scores when the salient object did contain holes.

Altogether, these findings highlight that a valuable step for tackling many of this population's VQA goals is to initially locate the salient object.  That is because the answer will likely only be grounded in the salient object \emph{or} the background rather than their intersection.

\section{Algorithm Benchmarking}
\label{algobench}

We finally analyze modern salient object detection algorithms to show how they perform on our new dataset. We conducted all experiments on a Nvidia A100 GPU.  

\subsection{Experimental Design}

\paragraph{Dataset Splits.}
We adhere to the standard splits adopted for all VizWiz-based datasets (e.g., ~\cite{gurari2020captioning}), which translates to approximately a 60/20/20 training, validation and test split for our VizWiz-SO dataset; i.e., 19,116, 6,105, and 6,779 images respectively. 

\vspace{-1em}\paragraph{Evaluation Metrics.}
We evaluate each model with respect to five popular metrics for salient object detection models: Mean Absolute Error $(MAE)$, Structure Measure $(S_m)$, Mean F-Measure $(F_m)$, Enhanced Alignment Measure $(E_m)$, and Intersection over Union $(IoU)$.

\begin{table*}[ht!]
\begin{tabular}{crcccccccccc} \cline{1-12}
 &  & HP & VST & PGNet & DIS & ICON & TRACER & IPR & IPR-FT & IPR-S & IPR-DS \\
 &  & & \cite{vst} & \cite{pgnet} & \cite{dis} & \cite{zhuge2021salient} & \cite{lee2022tracer} & \cite{kim2022revisiting} \\ \hline\hline
\multirow{3}{*}{\rotatebox[origin=c]{90}{Attr.}} & Training set & - & D & DH & DIS5K & D & D & DHU & VW & VW & DV \\ \cline{2-12}
 & Input size & - & $224^2$ & $224^2,1024^2$ & $1024^2$ & $352^2$ & $640^2$ & $384^2$ & $384^2$ & $384^2$ & $384^2$ \\ \cline{2-12}
 & Size (MB) & - & 171 & 280 & 169 & 251 & 630 & 351 & 351 & 351 & 351 \\ \cline{1-12}
\multirow{5}{*}{\rotatebox[origin=c]{90}{VizWiz-SO}} & $MAE$ ↓ & 0.02 & 0.17 & 0.21 & 0.36 & 0.20 & 0.21 & \textbf{0.03} & \textbf{0.03} & 0.04 & 0.04 \\ \cline{2-12}
 & $S_m$ ↑ & 0.92 & 0.65 & 0.62 & 0.46 & 0.69 & 0.67 & 0.90 & \textbf{0.91} & 0.89 & 0.90 \\ \cline{2-12}
 & $F_m$ ↑ & 0.96 & 0.83 & 0.79 & 0.61 & 0.86 & 0.83 & \textbf{0.95} & \textbf{0.95} & 0.94 & 0.94 \\ \cline{2-12}
 & $E_m$ ↑ & 0.97 & 0.76 & 0.74 & 0.55 & 0.73 & 0.65 & \textbf{0.94} & \textbf{0.94} & 0.93 & 0.94 \\ \cline{2-12}
 & $IoU$ ↑ & 0.94 & 0.73 & 0.67 & 0.49 & 0.70 & 0.68 & \textbf{0.93} & \textbf{0.93} & 0.90 & 0.91 \\ \cline{1-12}
\end{tabular}
\centering
\caption{Analysis of existing models that we benchmark on our VizWiz-SO dataset, including both off-the-shelf models as well as the top-performing algorithm fine-tuned (-FT), trained from scratch using VizWiz-SO (-S), and trained from scratch using both DUTS and VizWiz-SO (-DS). We first report differentiating attributes between the algorithmsand then present the model performance with respect to five widely-used metrics. (HP=Human Performance; IPR=InSPyReNet; DH=D+HR; DHU=D+HR+UH; VW=VizWiz-SO; DV=DUTS+VizWiz-SO; D=DUTS-TR~\cite{wang2017}; VW=VizWiz-SO; HR=HRSOD; UH=UHRSD~\cite{zeng2019towards})}
\label{tab:metric_table}
\end{table*}

\vspace{-1em}\paragraph{Algorithms.}
We benchmark the following ten methods from the past four years to assess the difficulty of our new dataset for modern salient object detection models: 

\begin{itemize}[itemsep=0.05em,leftmargin=*]
\item \textit{Boundary Aware Segmentation Network (BASNet)} \cite{basnet}, 2019: can run at 70fps, making it suitable for real-world applications like those motivated by our dataset.
\item \textit{Fusion, Feedback and Focus Network (F3Net)} \cite{f3net}, 2020: was state-of-the-art on five datasets.
\item \textit{U2 Network (U2Net)} \cite{u2net}, 2020: is compact (4.7MB), making it suitable for resource-constrained devices such as smartphones (i.e., the image source for our use case).
\item \textit{Visual Saliency Transformer (VST)} \cite{vst}, 2021: was state-of-the-art, with a purely transformer architecture. 
\item \textit{Pyramidal Feature Shrinking Network (PFSNet)} \cite{pfsnet}, 2021: was state-of-the-art on five datasets.
\item \textit{Pyramid Grafting Network (PGNet)} \cite{pgnet}, 2022: state-of-the-art on five datasets using a one-stage framework based on a transformer and CNN backbone. 
\item \textit{Dichotomous Image Segmentation (DIS)} \cite{dis}, 2022: state-of-the-art for related task of segmenting foreground objects (even if not salient) in high resolution images. 
\item \textit{Salient Object Detection via Integrity Learning (ICON)} \cite{zhuge2021salient}, 2022: state-of-the-art at time of publication.
\item \textit{Extreme Attention Guided Salient Object Tracing Network (TRACER)} \cite{lee2022tracer}, 2022: the current state-of-the-art model on DUTS challenge.
\item \textit{Revisiting Image Pyramid Structure for High Resolution Salient Object Detection (InSPyReNet)} \cite{kim2022revisiting}, 2022: the current $2^{nd}$ ranked model on DUTS challenge.
\end{itemize}

\noindent
We further characterize each model in Table \ref{tab:metric_table} by identifying datasets used for model training, image size used for model training, and model size.  All models predict segmentation maps showing pixel brightness, which we then convert into binary masks.  

\vspace{-1em}\paragraph{Humans.}
We evaluate human performance to establish an upper bound for automated methods by comparing the two annotations in cases where their IoU is greater than 0.90.\footnote{We chose this threshold to match what was used for establishing ground truth during dataset creation.}

\subsection{Performance for Off-The-Shelf Models}
We first evaluate all models in their original design. Key results are shown in Table \ref{tab:metric_table}, and the rest are in the Supplementary Materials due to space constraints. 

We observe that the top-performing model, InSPyReNet~\cite{kim2022revisiting}, nearly reaches human performance.  For example, the gap in MAE performance is $0.01$. Towards understanding why other models may perform far worse than human performance, we later explore whether it may be due to the training data or the architectures themselves.

We found that all models except InsPyReNet \cite{kim2022revisiting} perform worse on VizWiz-SO than on datasets which they were originally evaluated.  For example, the $MAE$ and $S_m$ performance observed by PGNet \cite{pgnet} on DUTS-TE is $0.028$ and $0.912$ respectively versus $0.2123$ and $0.6233$ respectively for our dataset.  These findings underscore that our dataset can offer distinct challenges from existing SOD datasets.  Analysis in subsequent sections reveal what makes our new dataset challenging for many modern models.

We also observe that the performance of methods designed for real-world, practical challenges, specifically to run fast (i.e, BASNet~\cite{basnet}) and be compact (i.e., U2Net~\cite{u2net}), are among the poorer-performing methods.  A tangentially important direction for future research will therefore be to ensure that models not only produce accurate saliency maps but also meet these practical needs.


\begin{table*}[ht!]
\centering
\begin{tabular}{llccccccccc}
\hline
 &  & VST & PGNet & DIS & ICON & TRACER & IPR & IPR-FT & IPR-S & IPR-DS \\
 &  & \cite{vst} & \cite{pgnet} & \cite{dis} & \cite{zhuge2021salient} & \cite{lee2022tracer} & \cite{kim2022revisiting} & \\ \hline\hline
\multirow{2}{*}{Text Present} & True & 0.13 & 0.16 & 0.32 & 0.15 & 0.18 & 0.02 & 0.02 & 0.03 & 0.03 \\ \cline{2-11} 
 & False & 0.24 & 0.29 & 0.40 & 0.24 & 0.25 & 0.09 & 0.09 & 0.10 & 0.11\\ \hline
\multirow{3}{*}{Coverage} & Small & 0.11 & 0.12 & 0.10 & 0.12 & 0.13 & 0.07 & 0.07 & 0.08& 0.09 \\ \cline{2-11} 
 & Medium & 0.09 & 0.15 & 0.25 & 0.10 & 0.13 & 0.05 & 0.05 & 0.06 & 0.07 \\ \cline{2-11} 
 & Large & 0.30 & 0.35 & 0.70 & 0.35 & 0.39 & 0.08 & 0.09 & 0.10 & 0.10 \\ \hline
\multirow{2}{*}{Boundary} & High & 0.12 & 0.16 & 0.21 & 0.17 & 0.14 & 0.05 & 0.05 & 0.06 & 0.07 \\ \cline{2-11} 
 & Low & 0.21 & 0.25 & 0.48 & 0.26 & 0.24 & 0.08 & 0.07 & 0.08 & 0.08 \\ \hline
 \multirow{2}{*}{Resolution} & High & 0.16 & 0.22 & 0.37 & 0.21 & 0.19 & 0.09 & 0.09 & 0.11 & 0.11 \\ \cline{2-11} 
 & Low & 0.17& 0.21 & 0.35 & 0.20 & 0.21 & 0.08 & 0.08 & 0.10 & 0.10 \\ \hline
\multirow{2}{*}{Quality} & Good & 0.14 & 0.17 & 0.30 & 0.16 & 0.19 & 0.04 & 0.03 & 0.05 & 0.06 \\ \cline{2-11} 
 & Poor & 0.27 & 0.34 & 0.50 & 0.33 & 0.32 & 0.10 & 0.10 & 0.11 & 0.11 \\ \hline
\end{tabular}
\centering
\vspace{0.1em}
\caption{Fine-grained analysis of existing models with respect to presence of text on the salient object (``Text Present"), relative size of the salient object in the image (``Coverage"), relative complexity of the salient object's boundary (``Boundary"), and image quality (``Quality") using the $MAE$ ↓ metric. As shown, the models perform worse when salient objects lack text, occupy a large portion of the image, and have less complex boundaries as well as when the image quality is poor. (IPR=InSPyReNet)}
\label{tab:fine-grained-tab}
\end{table*}

\subsection{Performance When Training on VizWiz-SO}
We now explore whether training on our new dataset boosts the performance for the top-performing algorithm, InSPyReNet~\cite{kim2022revisiting}.  We analyze three variants: (1) pretrained InSPyReNet model fine-tuned on VizWiz-SO (IPR-FT), (2) InSPyReNet algorithm trained from scratch on VizWiz-SO (IPR-S), and (3) InSPyReNet algorithm trained from scratch on DUTS~\cite{wang2017} and VizWiz-SO (IPR-DS).  Overall, we observe slightly worse or comparable performance from these variants, suggesting that the in-domain training data from VizWiz-SO is not necessary. Only leveraging the greater diversity of training data VizWiz-SO provides, by fine-tuning on Viz-Wiz-SO, yields a very slight advantage.

\subsection{Fine-grained Analysis}
We next analyse what makes our dataset challenging for modern algorithms.  To do so, we divide the test set according to the following \textit{five} factors, with the first four based on metadata from Section~\ref{dataset_analysis} to characterize our dataset:

\begin{itemize}[itemsep=0.05em,leftmargin=*]
\item \textit{Text Presence}: two groups based on whether text is present in the salient object.

\item \textit{Coverage Ratio (Coverage)}: three groups based on the $33^{rd}$ and $66^{th}$ quartile values in our dataset. All images with coverage ratio less than $0.32$ has \textit{small} coverage, between $0.32$ and $0.62$ has \textit{medium} coverage, and greater than $0.62$ has \textit{large} coverage.

\item \textit{Boundary Complexity (Boundary)}: two groups by splitting them around the mean score for boundary complexity (i.e., $0.66$) with \textit{high} complexity when the score is less than the mean and \textit{low} complexity otherwise.

\item \textit{Image Resolution (Resolution)}: two groups by splitting images around whether they are high resolution, as defined by whether the image width and height are both at least 1080 and 1920 respectively.

\item \textit{Quality}: leveraging metadata from prior work \cite{gurari2020captioning}, which indicates how many of the five crowdworkers indicated an image as insufficient quality to recognize the content, we split the images into groups with \textit{good} quality being when none of the crowdworkers indicate insufficient quality and \textit{poor} otherwise.
\end{itemize}

\noindent
\emph{To our knowledge, this paper is the first to systematically characterize SOD algorithm performance with respect to these criteria.}  We again split results with key ones shown in Table \ref{tab:fine-grained-tab} and the rest shown in the Supplementary Materials due to space constraints.

In terms of text presence, we see that the models perform better when there is text present as opposed to when there is none. For example, the performance drops by 0.11 for the second best model, VST, and 0.07 for the top-performing model InSPyReNet.  We suspect visual patterns that arise with text may serve as a valuable cue to models in locating salient objects.

Next, we see that as the coverage ratio of the salient objects increase, the models tend to perform worse with all models performing the worst on the large objects. For instance, the second best model, VST, has a performance dropoff of 0.19 when predicting images with small coverage ratios as opposed to large coverage ratios.  In contrast, the top-performing model InSPyReNet only has a drop of 0.01, which we suspect is a large reason behind this model's superior performance.  We suspect this performance gap arises in part from existing datasets largely lacking such large salient objects, which both could have affected what algorithms were designed to handle as well what they could learn from the data they observed.

Further observed trends are that performance drops for salient objects with lower boundary complexity and poorer quality images.  We suspect this is due to domain shifts between our dataset and prior datasets that affect what algorithms were designed for and what they could learn from the training data.
\section{Conclusions}
\label{conc}
This work's contributions are: (1) a new SOD dataset originating from a practical use case (Section~\ref{dataset-creation-process}), (2) comparison of this dataset to seven popular SOD datasets to reveal how it fills important gaps of existing SOD datasets (Section~\ref{sec_comparisonToOtherDatasets}), (3) comparison of this dataset to the VQA grounding dataset to reveal how it relates (Section~\ref{sec:SOvsVQA}), and (4) benchmarking of seven modern algorithms on this dataset to reveal limitations of modern SOD algorithms (Section~\ref{algobench}). To our knowledge, this work is the first to (1) systematically analyze SOD datasets based on whether text is present, holes are present, multiple regions are present, and their relationship to VQA tasks (Section~\ref{dataset_analysis}) and (2) systematically analyze SOD algorithms with respect to whether text is present, image quality, as well as levels of an object’s boundary complexity and image coverage ratio (Section~\ref{algobench}). Results reveal domain shifts between our SOD dataset and existing SOD datasets as well as limitations of many SOD algorithms: they struggle for salient objects that are large, with simpler boundaries, and lack text as well as for lower quality images. Results also show that one algorithm achieves nearly human performance.




\vspace{1em}
\begin{flushleft}
    \textbf{Acknowledgments.} 
    This project was supported in part by a National Science Foundation SaTC award (\#2148080) and Amazon Mechanical Turk. We thank Leah Findlater and Yang Wang for contributing to this research idea.
\end{flushleft}

\clearpage
{\small\bibliographystyle{ieee_fullname}\bibliography{egbib}}
\clearpage

\pagebreak
\section*{Appendix}
\noindent
This document supplements the main paper with additional information concerning:
\begin{enumerate}
    \item Dataset Creation (supplements Section 3.1)
    \begin{itemize}
        \item Annotation Task Interface 
        \item Worker Qualification Task
        \item Analysis of Workers' Annotation Differences 
    \end{itemize}
    \item Experimental Design (supplements Section  4.1)
    \item Experimental Results (supplements Sections  4.2-4.4)
\end{enumerate}

\renewcommand{\thesection}{\Alph{section}}
\setcounter{section}{0}

\section{Dataset Creation}
\subsection{Annotation Task Interface}
\begin{figure*}[t!]
    \centering
    \includegraphics[width=\textwidth]{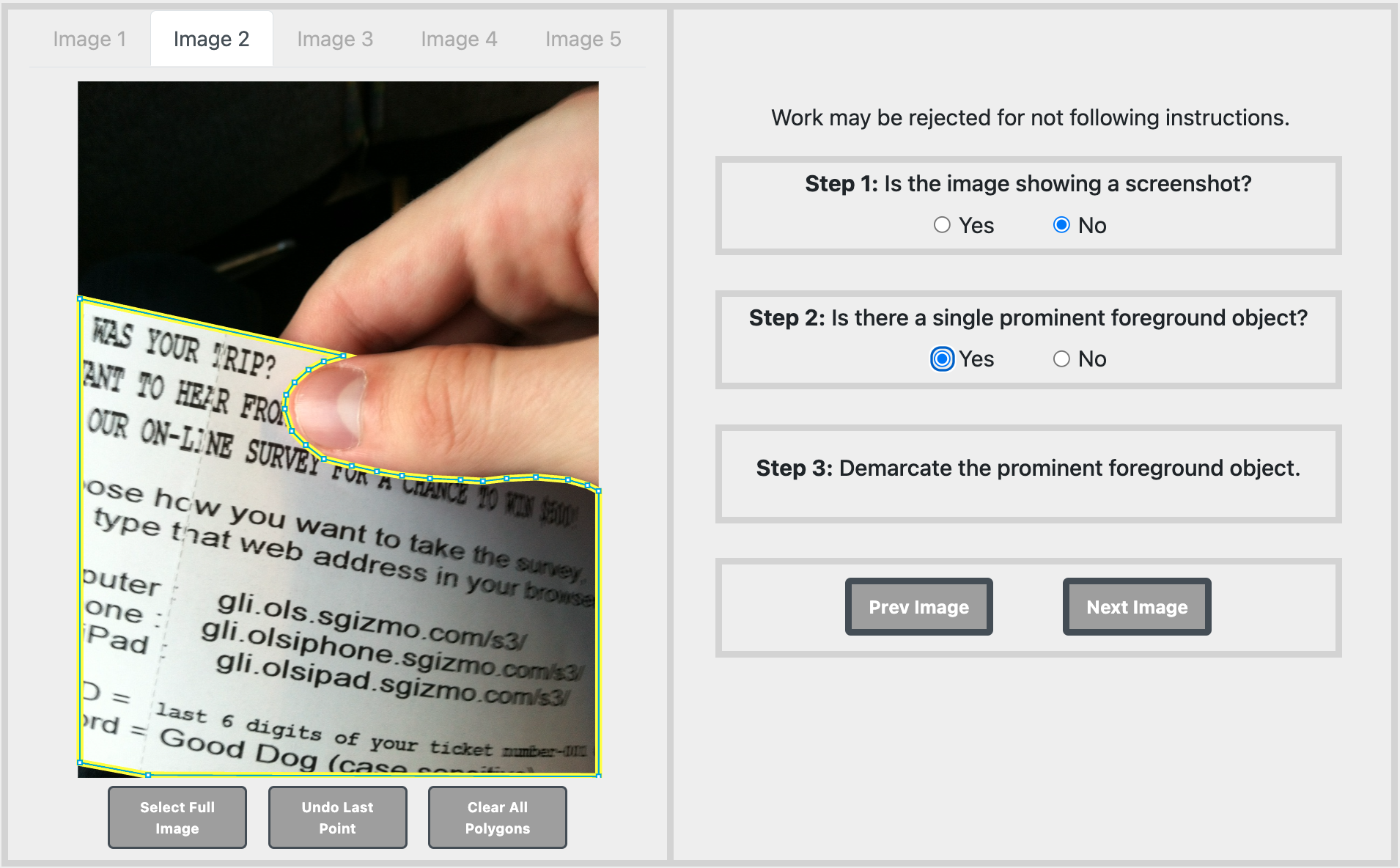}
    \caption{A screenshot of our annotation task interface.}
    \label{fig:task_interface}
    \centering
\end{figure*}

The task interface displays five images within a tabbed container on the left and preliminary questions with task instructions on the right. A screenshot of the task interface (without instructions) is shown in Figure \ref{fig:task_interface}.

To account for occlusions and holes while keeping the task simple for annotators, we permitted annotators to generate multiple polygons. For occlusions, annotators could use as many polygons as necessary for demarcating foreground objects partitioned into multiple polygons. For holes, we apply an even-odd fill rule to images featuring foreground objects with holes. With an even-odd fill rule, every area inside an even number of enclosed areas becomes hollow, and every region inside an odd number of enclosed areas becomes filled \cite{evenodd-fill-rule}. By treating the image's four corners as the first enclosed area, the outermost boundary of the foreground object becomes the second enclosed area. Moreover, holes within foreground objects represent the third layer of enclosed areas and become filled, allowing annotators to demarcate foreground objects featuring holes. In practice, annotators first trace the outermost boundary of the foreground object and close the path by clicking the first point a second time. We then instructed annotators to trace any holes within the foreground object, and so those holes end up in odd-numbered layers. 

\subsection{Worker Qualification Task}
\begin{figure}[t!]
    \begin{center}
    	\includegraphics[scale=1, width=24em]{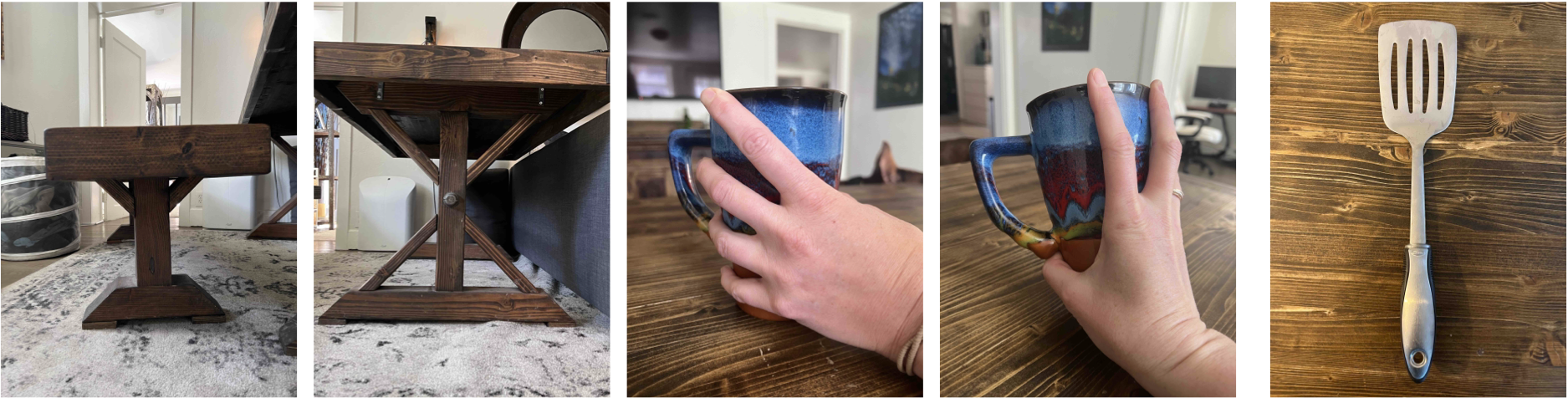}
    	\caption{The five images used for the worker qualification task. Each was selected to demonstrate a challenging annotation scenario such as complex boundaries, holes, and occlusions.}
    	\label{qual_images}
    \end{center}
\end{figure}

We administered a qualification task for workers to support our collection of high-quality ground truth annotations. The qualification task required annotating five images, each of which features a distinct challenging annotation scenario.  All five images are shown in Figure \ref{qual_images}.  The first two images show a table and a bench, offering examples with complex boundaries and holes.  The next two images feature a person holding a coffee mug, to support educating a crowdworker about our expectations for annotating objects with complex geometries that have many curves and occlusions that require annotating multiple polygons. The final image is a spatula. This task verified a crowdworker's ability to correctly identify and annotate multiple holes that can arise within the salient object.

After crowdworkers annotated each qualification image, the backend code of our website checked if their annotation was sufficiently similar to the GT annotation (i.e., IoU similarity of at least 0.90). Crowdworkers could only proceed to the following image after they obtained an IoU \begin{math}\geq\end{math} 0.90 on the current image. Crowdworkers obtaining an IoU \begin{math}\geq\end{math} 0.90 on all five qualification assessment images on a per-image basis gave us substantial confidence that they would be able to successfully handle complex and challenging outlier cases within the original VizWiz Dataset.\footnote{Some crowdworkers did not pass the qualification assessment due to time constraints. In these cases, crowdworkers would contact us with the images they annotated. If we were confident in their annotation abilities, we manually added these crowdworkers to the qualified worker pool.}

\begin{figure}[t!]
    \begin{center}
    	\includegraphics[scale=1, width=24em]{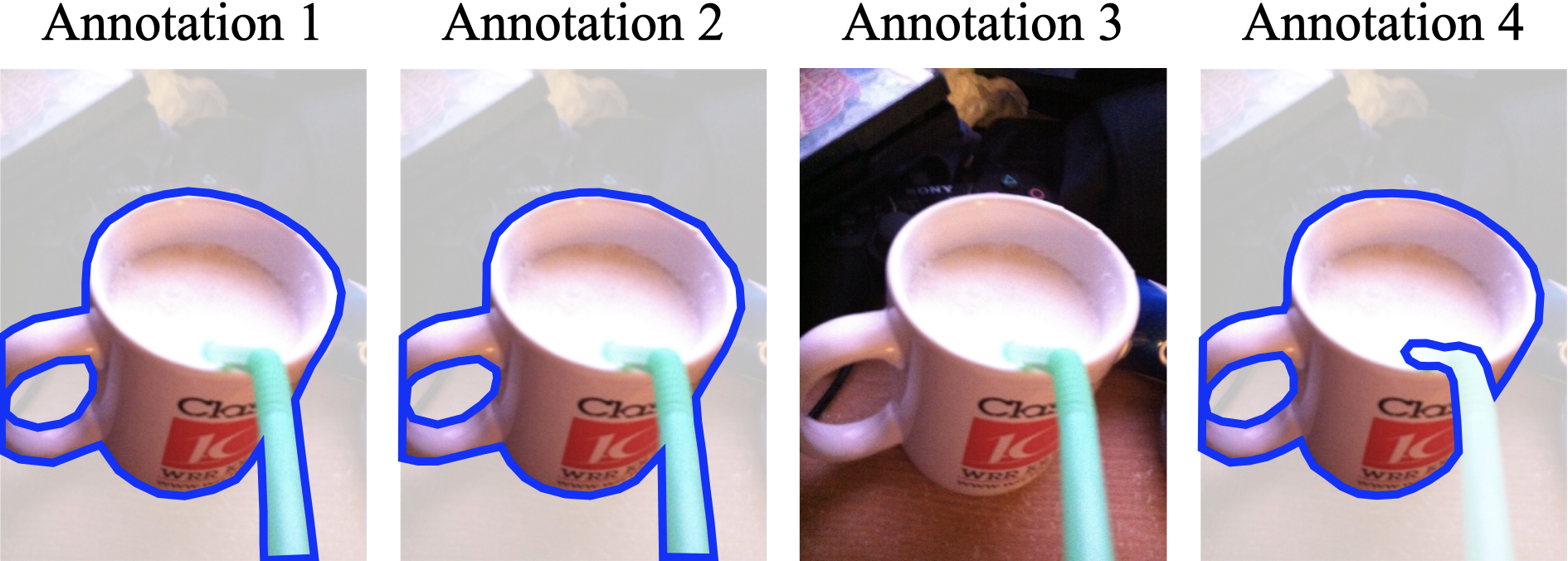}
    	\includegraphics[scale=1, width=24em]{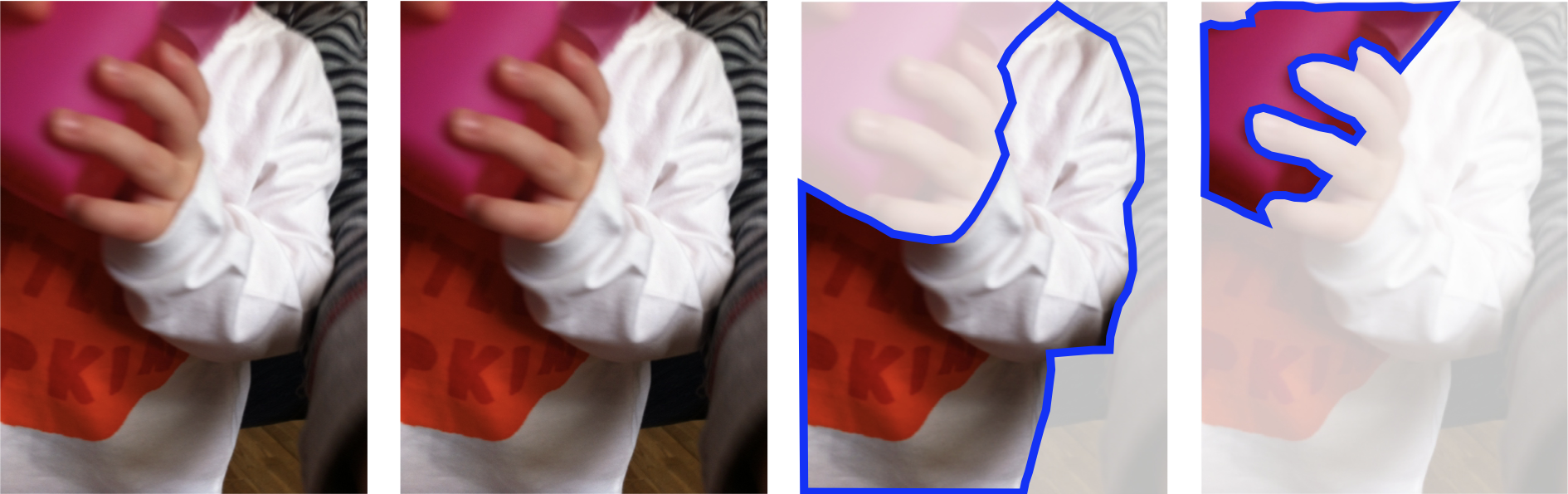}
    	\includegraphics[scale=1, width=24em]{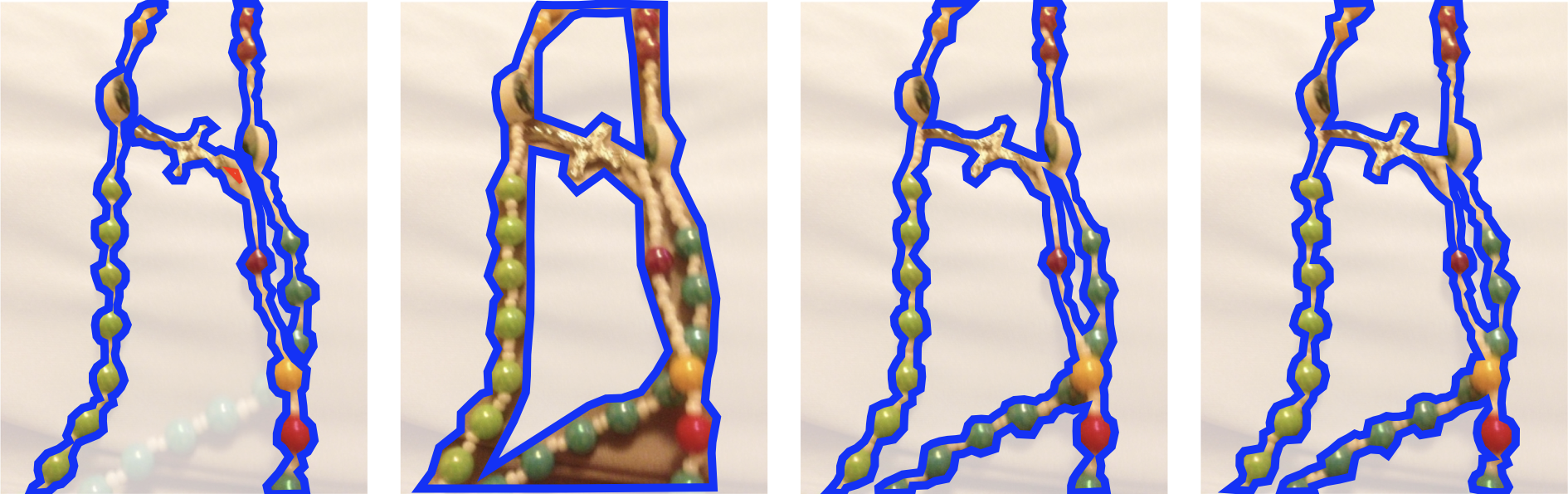}
    	\caption{Example annotations from our random subset where we collected four annotations as opposed to two. We find worker differences primarily occur in challenging annotation scenarios such as holes, occlusions, complex boundaries, and object saliency.}
    	\label{random_annotation}
    \end{center}
\end{figure}

\subsection{Analysis of Workers’ Annotation Differences}
We collected a larger number of redundant annotations per image for a random subset of images to better explore when and why annotation differences are observed from different workers. Specifically, for this analysis, we collected four annotations as opposed to two for a subset of 1,237 images. Examples of the redundant annotations collected per image are shown in Figure \ref{random_annotation}. 

The first example (i.e., row 1 of Figure \ref{random_annotation}) highlights that annotation differences can stem from challenging annotation scenarios where objects contain holes (e.g., in mug handle) or are occluded (e.g., by the straw).  For instance, the hole was not annotated in the third annotation.  Additionally, only the fourth annotation captured the occlusion that arises from the straw.

The second example (i.e., row 2 of Figure \ref{random_annotation}) highlights that annotation differences can stem from ambiguity regarding what is the salient object.  As shown, the first two annotations flag the image as lacking a foreground object, the third annotation identifies the child holding the cup as the salient object, and the fourth annotation identified the child's cup as the salient object.

The third example (i.e., in row 3 of Figure \ref{random_annotation}) highlights that annotation differences also can arise for objects that simultaneously have complex boundaries and holes. In annotation one, the worker did not fully annotate the salient object, cutting out part of the object from the annotation. Only the third and fourth annotations accurately annotate all holes that are present in the salient object's boundary while also having tight boundaries in the annotation.


In summary, we found occlusions, holes, and saliency ambiguity to be the primary factors contributing to annotation differences. In the case of occlusions, worker differences can arise when deciding whether to include objects that are a composite part of the salient object. In the case of holes, annotation differences can arise regarding which holes to annotate. Last, we found that it can be ambiguous as to which object is the most salient.  To facilitate future analysis of human performance, we will publicly share metadata with all humans’ annotations for all images.

\section{Experimental Design}
We compute the five metrics used in the benchmarking section using the following definitions:

\textit{Mean Absolute Error} \cite{MAE} represents the average absolute difference between the predicted saliency map and its ground truth per pixel. It can be given as:
\begin{align}
    MAE = \frac{1}{H*W}\sum^H_{r=1}\sum^W_{c=1}|pred(r, c) - gt(r, c)|
    \label{mae}
\end{align}
where $pred$ represents the predicted saliency map, $gt$ represents the ground truth, $(H, W)$ represents the height and width of the image, and $(r, c)$ represents the pixel co-ordinates for the given image.

\textit{Structure Measure} \cite{Smeasure} is used to measure the similarity between the predicted saliency map and the ground truth. Since, we convert both the predictions and ground truths into the $[0, 1]$ range, we apply the formula directly to the predictions and maps. It can defined as follows:
\begin{align}
    S_m = (1 - \alpha) S_r + \alpha S_o
\end{align}
where, $S_r$ is defined as the region aware similarity score, $S_o$ is defined as the object aware similarity score, and $\alpha$ represents the weight that is used to sum up the values. We set $\alpha = 0.5$, therefore making sure that we see equal contribution from both region and object aware scores.

\textit{F-Measure} \cite{Fmeasure} represents the precision and recall ratio for the given prediction. It can be represented as:
\begin{align}
    F_m = \frac{(1 + \beta^2) * Precision * Recall}{\beta^2 * Precision + Recall}
\end{align}
Here $precision = \frac{TP}{TP + FP}$ and $recall = \frac{TP}{TP + FN}$ on the entire prediction image by pixels. We set $\beta^2 = 0.3$ and report the average of all F-measures as $F_m$ similar to previous works.

\textit{Enhanced Alignment Measure} \cite{Emeasure} is used as the metric to measure the effectiveness of the saliency prediction against the ground truth. It captures the pixel-level matching information and image-level statistics into one single metric by the means of an enhanced alignment matrix $\phi$. It is defined as follows:
\begin{align}
    E_m = \frac{1}{H*W}\sum^H_{r=1}\sum^W_{c=1}\phi_{FM}(r, c)
\end{align}
where, $\phi_{FM}$ represents the enhanced alignment matrix for the foreground map, $(H, W)$ represents the height and width of the image, and $(r, c)$ represents the pixel co-ordinates for the given image.

\textit{Intersection over Union} also known as Jaccard Index is used to determine the similarity between sample sets. In this case it captures the overlap between the ground truth and prediction map of the salient object. We convert the predictions in binary map and compute the Jaccard Index over two classes. It can be defined as follows:
\begin{align}
    IoU = J(A, B) = \frac{|A \cap B|}{|A \cup B|}
\end{align}
where, $A$ and $B$ are images of same size, consisting of integer class values $\{0, 1\}$.

\section{Algorithm Benchmarking}
We provide more details about our algorithm benchmarking here.  First, we report each model's backbone in Table \ref{tab:training_backbone_new}. Second, we show results for SOD models mentioned in the paper that are older.  Of note, for fine-tuning the InSPyReNet model and training the InSPyReNet model from scratch using VizWiz-SO and DUTS+VizWiz-SO, we modify the training hyperparameters to fit the GPU requirements available to us. Specifically, we reduce the \texttt{batchsize} to 4, \texttt{num\_worker} to 4, \texttt{epochs} to 40, and \texttt{warmup\_iterations} to 1000.  We also report results for three variants of the second-best model, VST~\cite{u2net}: (1) pretrained model fine-tuned on VizWiz-SO (VST-FT), (2) algorithm trained from scratch on VizWiz-SO (VST-S), and (3) algorithm trained from scratch on DUTS~\cite{wang2017} and VizWiz-SO (VST-DS).   Overall results are shown in Table \ref{tab:training_opt_old} and fine-grained analysis of these models are shown in Table \ref{tab:fine-grained-tab_old}.

\begin{table*}[ht!]
\centering
\begin{tabular}{rccccccc} \cline{1-8}
 & HP & VST & PGNet & DIS & ICON & TRACER & IPR \\
 & & \cite{vst} & \cite{pgnet} & \cite{dis} & \cite{zhuge2021salient} & \cite{lee2022tracer} & \cite{kim2022revisiting} \\ \hline\hline
Backbone & - & T2T-ViT & R-18+Swin & U2Net & Swin & ENet-7 & Swin  \\
\hline
\end{tabular}
\caption{Details of the various backbones used by the algorithms used for benchmarking. (ViT=Vision Transformer~\cite{vit}; R=ResNet~\cite{resnet};  Swin=Shifted window transformer~\cite{swin}; ENet=EfficientNet~\cite{tan2020efficientnet})}
\label{tab:training_backbone_new}
\centering
\end{table*}

\begin{table*}[ht!]
\centering
\begin{tabular}{crccccccc}
\hline
 &  & BASNet & F3Net & U2Net & PFSNet & VST-FT & VST-S & VST-DS \\
 &  & \cite{basnet} & \cite{f3net} & \cite{u2net} & \cite{pfsnet} & & & \\ \hline\hline
\multirow{3}{*}{\rotatebox[origin=c]{90}{Attr.}} & Backbone & R-34 & R-50 & - & R-50 & ViT & ViT& ViT \\ \cline{2-9} 
 & Training set & D & D & D & VW & VW & D+VW & D \\ \cline{2-9} 
 & Input size & $256^2$ & $352^2$ & $320^2$ & $352^2$ & $224^2$ & $224^2$ & $224^2$ \\ \cline{2-9} 
 & Size (MB) & 333 & 98 & 4.7 & 120 & 171 & 171 & 171 \\ \hline
\multirow{5}{*}{\rotatebox[origin=c]{90}{VizWiz-SO}} & $MAE$ ↓ & 0.28 & 0.28 & 0.26 & 0.32 & 0.19 & 0.21 & 0.23 \\ \cline{2-9} 
 & $S_m$ ↑ & 0.59 & 0.55 & 0.61 & 0.48 & 0.64 & 0.63 & 0.58 \\ \cline{2-9} 
 & $F_m$ ↑ & 0.77 & 0.74 & 0.80 & 0.70 & 0.74 & 0.72 & 0.68 \\ \cline{2-9} 
 & $E_m$ ↑ & 0.64 & 0.65 & 0.65 & 0.60 & 0.77 & 0.70 & 0.70 \\ \cline{2-9} 
 & $IoU$ ↑ & 0.62 & 0.53 & 0.63 & 0.48 & 0.70 & 0.69 & 0.64 \\ \hline
\end{tabular}
\caption{Quantitative comparison of off-the-shelf models (which are cited) as well as the VST model after being fine-tuned (-FT), trained from scratch on our VizWiz-SO dataset (-S), and trained from scratch on both DUTS and VizWiz-SO datasets (-DS). D=DUTS-TR \cite{wang2017}; VW=VizWiz-SO; R=ResNet\cite{resnet}; VST=Visual Saliency Transformer\cite{vst}; ViT=Vision Transformer\cite{vit}}
\label{tab:training_opt_old}
\centering
\end{table*}

\begin{table*}[ht!]
\centering
\begin{tabular}{llccccccc}
\hline
 &  & BASNet & F3Net & U2Net & PFSNet & VST-FT & VST-S & VST-DS \\
 &  & \cite{basnet} & \cite{f3net} & \cite{u2net} & \cite{pfsnet} & & & \\ \hline\hline
\multirow{2}{*}{Text Present} & True & 0.23 & 0.22 & 0.22 & 0.25 & 0.16 & 0.17 & 0.18 \\ \cline{2-9} 
 & False & 0.35 & 0.38 & 0.32 & 0.42 & 0.24 & 0.26 & 0.28 \\ \hline
\multirow{3}{*}{Coverage} & Small & 0.06 & 0.16 & 0.07 & 0.16 & 0.09 & 0.11 & 0.14 \\ \cline{2-9} 
 & Medium & 0.15 & 0.20 & 0.15 & 0.24 & 0.09 & 0.10 & 0.11 \\ \cline{2-9} 
 & Large & 0.60 & 0.47 & 0.54 & 0.54 & 0.38 & 0.39 & 0.40 \\ \hline
\multirow{2}{*}{Boundary} & High & 0.15 & 0.21 & 0.15 & 0.24 & 0.11 & 0.12 & 0.12 \\ \cline{2-9} 
 & Low & 0.38 & 0.34 & 0.35 & 0.38 & 0.26 & 0.27 & 0.28 \\ \hline
 \multirow{2}{*}{Resolution} & High & 0.30 & 0.30 & 0.28 & 0.33 & 0.17 & 0.18 & 0.19 \\ \cline{2-9} 
 & Low & 0.26 & 0.27 & 0.26 & 0.31 & 0.19 & 0.20 & 0.21 \\ \hline
\multirow{2}{*}{Quality} & Good & 0.22 & 0.23 & 0.21 & 0.26 & 0.16 & 0.17 & 0.19 \\ \cline{2-9} 
 & Poor & 0.44 & 0.43 & 0.41 & 0.47 & 0.30 & 0.31 & 0.33 \\ \hline
\end{tabular}
\centering
\vspace{0.1em}
\caption{Fine-grained analysis of off-the-shelf models (which are cited) as well as the VST model after being fine-tuned (-FT), trained from scratch on our VizWiz-SO dataset (-S), and trained from scratch on both DUTS and VizWiz-SO datasets (-DS).  This covers analysis with respect to presence of text on the salient object (``Text Present"), relative size of the salient object in the image (``Coverage"), relative complexity of the salient object's boundary (``Boundary"), and image quality (``Quality") using the $MAE$ ↓ metric. As shown, the models perform worse when salient objects lack text, occupy a large portion of the image, and have less complex boundaries as well as when the image quality is poor.}
\label{tab:fine-grained-tab_old}
\end{table*}

\begin{figure*}[t]
    \centering
    \includegraphics[width=1\textwidth]{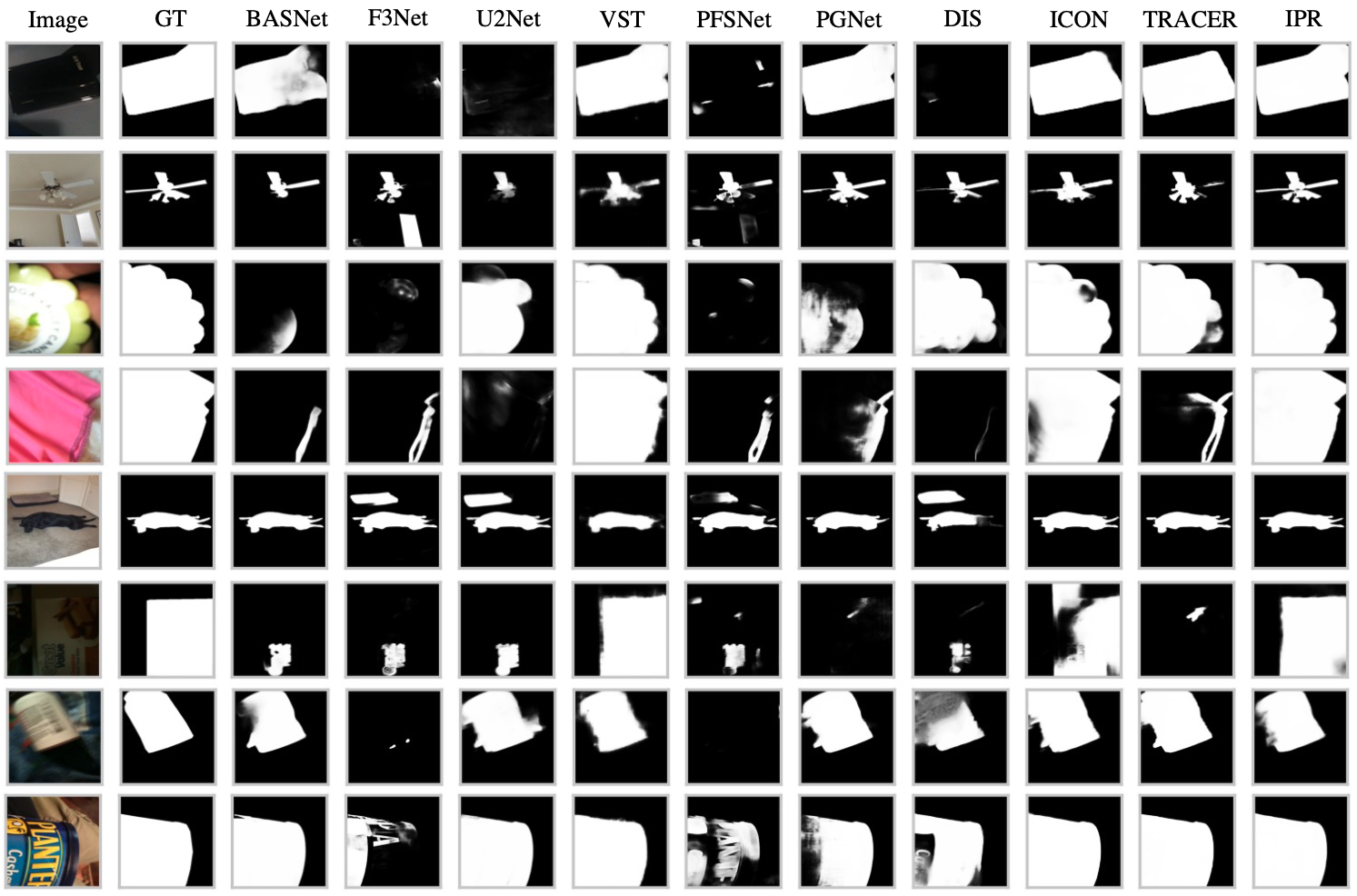}
    \caption{Examples of images with characteristics such as high coverage ratio, presence of text, less complex boundaries, and lower image quality. We show how the ten models perform on these cases as compared to the human annotation (GT=Ground Truth). We see that models such as PFSNet \cite{pfsnet}, DIS \cite{dis}, and F3Net \cite{f3net} do not always give us the correct salient objects or sometime no predictions at all. We also notice that VST \cite{vst}, ICON \cite{zhuge2021salient}, and InSPyReNet~\cite{kim2022revisiting} usually predicts salient objects with better accuracy compared to other models. InSPyReNet is the closest to human performance across all the various images presented here. (HP=Human Performance, IPR=InSPyReNet)}
    \label{fig:preds}
    \centering
\end{figure*}

\begin{figure*}[t]
    \centering
    \includegraphics[width=1\textwidth]{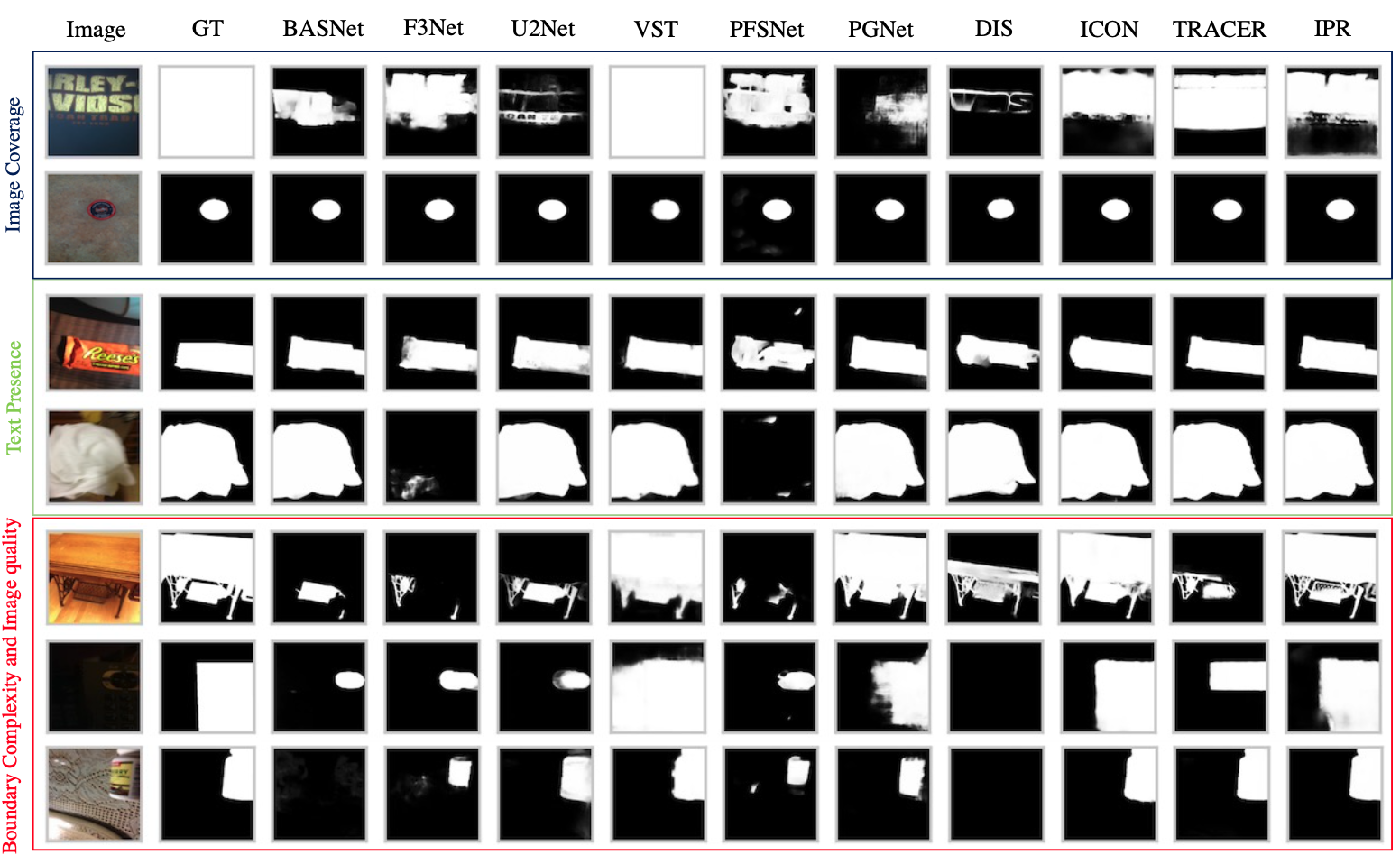}
    \caption{Examples of images with characteristics with respect to our fine-grained analysis that we found are challenging for modern salient object detection models. We show how the ten models perform on these cases as compared to the human annotations (GT=Ground Truth). We see that all models except InSPyReNet~\cite{kim2022revisiting} seem to perform poorly for objects with high coverage ratio and well for objects with lower coverage ratio. Next, we see for presence of text, that models do not perform worse when there is no text on the salient object. Finally, we see that all models except InSPyReNet~\cite{kim2022revisiting} can suffer from not detecting the correct salient object for salient objects with complex boundaries and for images with quality issues. Overall, InSPyReNet~\cite{kim2022revisiting} is the closest to the human performance. (IPR=InSPyReNet)}
    \label{fig:preds-2}
    \centering
\end{figure*}

We show qualitative examples for these models on VizWiz-SO in Figures \ref{fig:preds} and \ref{fig:preds-2}. These examples feature a variety of challenges we observed for the models in our fine-grained analysis. Most models perform poorly in identifying larger salient objects (rows 4 and 5 in Figure \ref{fig:preds} and row 1 in Figure \ref{fig:preds-2}), but perform relatively well on images with smaller salient objects (row 2 in Figure \ref{fig:preds-2}). We also observe the most models perform better when salient objects contain text (rows 1 and 2 in Figure \ref{fig:preds} and row 3 in Figure \ref{fig:preds-2}) versus lack text (rows 5 and 6 in Figure \ref{fig:preds} and row 4 in Figure \ref{fig:preds-2}). Further, we see most models perform worse for images with complex boundaries (row 5 in Figure \ref{fig:preds-2}) and that are lower quality (rows 3, 4, and 5 in Figure  \ref{fig:preds} and rows 6 and 7 in Figure \ref{fig:preds-2}).  

\end{document}